\documentclass[sigconf]{acmart}
\usepackage{listings}
\usepackage{xcolor} 
\usepackage{enumitem}
\setlist[itemize]{leftmargin=*}

\setcopyright{acmlicensed}
\copyrightyear{2026}
\acmYear{2026}
\acmDOI{XXXXXXX.XXXXXXX}
\acmConference[AGENT 2026]{1st International Workshop on Agentic Engineering}{April 14, 2026}{Rio de Janeiro, Brazil}
\acmISBN{978-1-4503-XXXX-X/2018/06}

\begin{document}

\title{AgentFixer: From Failure Detection to Fix Recommendations in LLM Agentic Systems}

\author{Hadar Mulian}
\affiliation{%
  \institution{IBM Research}
  \country{Israel}
  }
\email{hadar.mulian@ibm.com}
\author{Sergey Zeltyn}
\affiliation{%
  \institution{IBM Research}
  \country{Israel}
  }
\email{sergey.zeltyn@ibm.com}

\author{Ido Levy}
\affiliation{%
  \institution{IBM Research}
  \country{Israel}
  }
\email{ido.levy1@ibm.com}

\author{Liane Galanti}
\affiliation{%
  \institution{IBM Research}
  \country{Israel}
  }
  \email{liane.galanti@ibm.com}

\author{Avi Yaeli}
\affiliation{%
  \institution{IBM Research}
  \country{Israel}
  }
\email{avi.yaeli@ibm.com}

\author{Segev Shlomov}
\affiliation{%
  \institution{IBM Research}
  \country{Israel}
  }
\email{segev.shlomov1@ibm.com}
\renewcommand{\shortauthors}{Mulian et al.}

\begin{abstract}
We introduce a comprehensive validation framework for LLM-based agentic systems that provides systematic diagnosis and improvement of reliability failures. The framework includes fifteen failure-detection tools and two root-cause analysis modules that jointly uncover weaknesses across input handling, prompt design, and output generation. It integrates lightweight rule-based checks with LLM-as-a-judge assessments to support structured incident detection, classification, and repair. We applied the framework to IBM CUGA, evaluating its performance on the AppWorld and WebArena benchmarks. The analysis revealed recurrent planner misalignments, schema violations, brittle prompt dependencies, and more. Based on these insights, we refined both prompting and coding strategies, maintaining CUGA’s benchmark results while enabling mid-sized models such as Llama 4 and Mistral Medium to achieve notable accuracy gains, substantially narrowing the gap with frontier models. Beyond quantitative validation, we conducted an exploratory study that fed the framework’s diagnostic outputs and agent description into an LLM for self-reflection and prioritization. This interactive analysis produced actionable insights on recurring failure patterns and focus areas for improvement, demonstrating how validation itself can evolve into an agentic, dialogue-driven process. These results show a path toward scalable, quality assurance, and adaptive validation in production agentic systems, offering a foundation for more robust, interpretable, and self-improving agentic architectures.
\end{abstract}




\maketitle

\section{Introduction}
LLM-based agentic systems are rapidly moving from prototypes into production, where they orchestrate complex workflows involving planning, reasoning, and tool use. In enterprise and consumer contexts alike, organizations face competing pressures: the need for rapid deployment, operational safety, and compliance on one side, and cost, latency, and privacy constraints on the other. These pressures often favor the use of smaller, non-frontier models, which offer attractive efficiency and deployment advantages. Yet these models are also far more fragile, especially when judged by operational reliability rather than raw reasoning ability.

A primary barrier to dependable deployment arises from \emph{output reliability errors}-situations where the model's responses cannot be consumed by downstream components. We showed that parsing-related incidents account for nearly 38\% of all observed task failures in production agentic systems, making them the single largest source of execution breakdowns. Typical examples include malformed JSON, missing schema fields, and instruction non-compliance. Even minor deviations from expected formats can halt pipelines, trigger cascading failures across agents, and force costly human debugging. Such incidents erode user trust and undermine otherwise strong task-level reasoning capabilities.

Industry responses to these challenges are often ad hoc: brittle regex filters, late-stage schema enforcement, or one-off prompt tweaks that address specific failures but do not generalize. These practices slow iteration cycles, introduce technical debt, and do little to prevent future incidents. What is lacking are systematic, deployment-ready tools that proactively diagnose and mitigate failure modes before they escalate into production incidents. This operational gap-what we frame as an \emph{AgentOps challenge}-is increasingly critical as organizations adopt multi-agent systems for high-value applications.

To address this gap, we built a comprehensive agentic-system validation suite of fifteen tools spanning three stages:
\begin{itemize} \item \textbf{Prompt analysis:} tools for contradiction detection, coverage gaps, and edge-case handling in system prompts. \item \textbf{Input validation:} schema and format checks to ensure well-formed requests before execution. \item \textbf{Output validation:} tools for factual consistency, syntactic correctness of code, anomaly detection, and reasoning--action alignment.
\end{itemize}
Beyond stage-specific validation, several tools check for critical \emph{cross-stage consistencies}: alignment between prompts and inputs, coherence between prompt instructions and generated outputs, and end-to-end pipeline integrity. Three of the tools are lightweight and rule-based for efficiency, while the remainder use LLM-as-a-judge methods to capture complex semantic failures. Together, they provide both deterministic reliability and flexible semantic assessment, yielding a principled incident analysis and prevention framework.

We deployed this framework on \textbf{IBM CUGA} \cite{shlomov2025benchmarksbusinessimpactdeploying}, an enterprise-ready multi-agent system. Using 24 AppWorld task templates, the tools surfaced systemic weaknesses such as controller indexing errors, planner misalignments, and schema non-compliance. These findings form a taxonomy of recurrent error categories that directly hinder pipeline reliability. Motivated by this analysis, we refined CUGA's prompting strategies for its Web deployment. The revised prompts preserved its first-place ranking on WebArena and, critically, allowed mid-sized open-source models to achieve results competitive with GPT-based systems. This represents not only a research insight but also a deployment-level improvement: robust execution at lower cost and latency.

Our contributions are threefold: (1) a deployed validation tool framework that systematizes incident analysis for agentic systems, (2) a case study of IBM CUGA on AppWorld tasks that exposes recurrent error patterns and root causes, and (3) deployment improvements on WebArena and AppWorld that demonstrate how validation-driven refinement enables mid-sized models to achieve near frontier-level reliability.

\section{Related Work}

\label{sec:related_work}

Early work on dependable LLM outputs focused on local syntactic controls. Function calling and schema-guided prompting narrowed generations to typed JSON fields, while grammar constrained decoding guaranteed token-level validity, and community libraries generalized these ideas into portable parsing and type-checking layers \citep{schick2024jsonformer,pydanticai2024}. At scale, providers introduced structured outputs with cross-runtime schema fidelity guarantees \citep{openai2024structured,hf2024structured}. In parallel, post-generation validation-and-repair loops emerged: rule-based checkers triggered retries or localized fixes, and learned formatters transformed free text into schema-correct structures with near-perfect syntactic accuracy \citep{tam2024structured,wang2025slot,oved2025snap}. While effective at reducing parser failures, these approaches operate at the level of individual calls and can constrain expressivity or long-horizon reasoning in compositional tasks, motivating training-time methods and evaluation beyond syntactic correctness \citep{tam2024structured,zhang2025rljson}.

This recognition shifted attention from per-call guarantees to system-level reliability in multi-agent systems (MAS) \cite{marreed2025towards}, where correctness depends on how multiple decisions, tool calls, and messages compose over time. Recent frameworks characterize agentic failure modes: MAST organizes errors into specification/design issues, inter-agent misalignment, and verification failures \citep{cemri2025multi}, while TRAIL provides trajectory-level traces with labeled reasoning, planning, execution, and tool-use errors, on which even strong long-context models achieve only $\sim$11\% accuracy \citep{deshpande2025trail}.  Empirical studies further expose cross-agent pathologies—such as robustness under faulty participants and recurring root causes in incident analyses—highlighting the need for validators that reason over plans, handoffs, and shared state \citep{huang2024faulty,shlomov2024grounding}. Reviews and practice-oriented reports echo these themes, mapping production incidents to coordination and context-propagation risks \citep{he2025lmaSE,mohammadi2025survey,schwartz2023enhancing}. Collectively, this body of work provides diagnostic taxonomies and datasets for post-incident analysis, implicitly defining the invariants—plan–action consistency, tool pre/postconditions, and cross-agent state coherence—that runtime validators must enforce to prevent cascading failures.

Maintaining complex systems at scale has driven the adoption of production observability stacks. Open-source and commercial platforms provide tracing, latency and token analytics, and dataset-driven evaluation integrated with orchestration frameworks \citep{langsmith2024,moshkovich2025taming,zeltyn2022prescriptive}. In parallel, OpenTelemetry gained LLM- and agent-specific spans and attributes to standardize telemetry across heterogeneous providers \citep{openllmetry2025}, which vendors have incorporated into offerings with end-to-end chain and agent tracing, quality checks, and experiment management. \cite{phoenix2024review} provides an OTel-native workflow for evaluation and troubleshooting. A key divide is architectural awareness: framework-native tools (e.g., LangSmith, AgentOps, AutoGen Studio, CrewAI) enable deep, specialized introspection but risk lock-in \citep{autogenstudio2024paper,crewaiTesting2025}, whereas OTel-based streams and neutral evaluation stores favor portability at the cost of framework-specific detail \citep{openllmetry2025}.
Comparative analyses echo this trade-off, highlighting portability challenges for organizations that mix or migrate agent stacks \citep{langchain2025thinkabout}. Overall, observability explains what happened; the open problem is deciding whether it should have happened—and enforcing that decision during execution.

Automated semantic evaluators, ``LLM-as-a-Judge'' (LaaJ), approximate human grading at lower cost and show strong alignment with human preferences \citep{zheng2023judging,gu2025judge}. Research has also examined systematic biases and calibration strategies, including prompt design, aggregation methods, and mixed-initiative criteria selection \citep{shi2024positionbias,shankar2024evalgen}. While these advances enable scalable semantic assessment, most judge pipelines operate offline or at component boundaries rather than within online control loops \citep{zheng2023mtbench,openai_evals}. Meanwhile, structured-output and policy checks are typically enforced as separate layers for schemas, pre/postconditions, and safety \citep{guardrails_structured_2025}, and multi-agent failure taxonomies highlight cross-agent invariants that judges alone do not capture \citep{cemri2025multi,deshpande2025trail}. A \emph{trajectory-level} validator thus integrates deterministic checks with learned judges for semantic properties and applies both across multi-turn interactions instead of isolated calls \citep{zheng2023mtbench,liuagentbench}.

Taken together, the literature traces a clear arc: local syntactic controls improved parsability \citep{openai2024structured,schick2024jsonformer,wang2025slot,hf2024structured}; MAS taxonomies clarified where and why coordinated systems fail and supplied data for analysis \citep{cemri2025multi,deshpande2025trail,huang2024faulty}; observability standardized what happened while splitting into architecture-aware versus architecture-agnostic paradigms \citep{langfuse2024,langsmith2024,openllmetry2025,phoenix2024review}, and LaaJ offered scalable semantic assessment with growing attention to bias and calibration \citep{zheng2023judging,gu2025judge,shi2024positionbias}. Yet these strands matured largely in parallel: structured-output methods seldom reason about cross-agent state, taxonomies describe failures more than they prevent them, observability surfaces signals without prescribing policy (and often ties to specific frameworks), and judges assess content without binding it to tool-use constraints or plan consistency. 

This motivates the missing operational layer: an architecture-agnostic validation plane that (i) encodes taxonomy-informed policies, (ii) composes deterministic and semantic checks across input–prompt–output boundaries and agent handoffs, and (iii) integrates with standard telemetry to enact real-time enforcement. Concretely, such a layer (A) rejects a tool call whose preconditions are unsatisfied even if local JSON is valid, (B) flags plan–action drift when a sub-agent deviates from a role specification despite high judge scores, and (C) halts handoff when cross-agent state invariants (e.g., resource locks, safety constraints) are violated-aligning with production deployment goals behind AgentFixer’s fifteen diagnostic tools and two root-cause analyzers.

\section{Failure Taxonomy and Tool Design}
Our validation framework targets failure categories identified through systematic analysis of multi-agent system breakdowns. Drawing from TRAIL's execution tracing methodology and established taxonomies of multi-agent failures, we identified three primary failure dimensions: (i) input-output format compliance, (ii) cross-stage information consistency, and (iii) semantic reasoning-action alignment. We augmented these literature-derived categories with failure modes observed in production deployments of IBM CUGA, particularly focusing on parsing errors that account for 38\% of task failures. Less critical failure types (e.g., minor stylistic violations) were excluded to maintain focus on deployment-blocking issues.

\subsubsection{Validation Tools}
To operationalize these hybrid validation strategies, we developed a comprehensive suite targeting the specific failure modes observed in multi-agent systems. The suite comprises fifteen tools organized by the artifacts they analyze: (i) user inputs against system prompts, (ii) system prompts alone, (iii) generated outputs against system prompts, (iv) token-level anomalies in generated outputs, (v) code-specific validation for Python, and (vi) cross-stage information consistency. Each tool is implemented either as a lightweight rule-based method (e.g., regex, fuzzy matching, AST parsing) or as an LLM-based evaluator.

\paragraph{(i) Input-Prompt Compliance.}
These tools check user inputs against requirements stated in the system prompt prior to execution.
\begin{itemize}
\item \textbf{Input-Schema-Non-compliance-Detector} (LLM-based): verifies that user inputs satisfy the schema specified in the prompt. \emph{Input:} system prompt, user input. \emph{Output:} binary violation indicator.
\item \textbf{Input-Instructions-Non-compliance-Detector} (LLM-based): checks compliance of user inputs to the instructions stated in the prompt. \emph{Input:} system prompt, user input. \emph{Output:} binary violation indicator.
\item \textbf{Input-Format-Violation-Detector} (LLM-based): validates that user inputs follow the required format constraints. \emph{Input:} system prompt, user input. \emph{Output:} binary violation indicator.
\end{itemize}

\paragraph{(ii) System Prompt Analysis.}
These tools assess prompt quality and completeness.
\begin{itemize}
\item \textbf{Prompt-Consistency-Validator: internal contradictions} (LLM-based): detects cases where the system prompt contradicts itself, such as providing an instruction in one part of the prompt and negating it in another. \emph{Input:} system prompt. \emph{Output:} binary indicator (contradiction present or not).  

\item \textbf{Prompt-Consistency-Validator: example misalignment} (LLM-based): detects inconsistencies between the few-shot examples and the described task or expected output, for instance when examples fail to fully match the specified format or logic. \emph{Input:} system prompt with few-shot examples. \emph{Output:} binary indicator (misalignment present or not).  

\item \textbf{Edge-Case-Instruction-Validator} (LLM-based): evaluates whether instructions anticipate reasonable edge cases. \emph{Input:} system prompt. \emph{Output:} binary indicator for missing edge-case coverage.
\item \textbf{Few-Shot-Coverage-Validator} (LLM-based): analyzes the diversity and coverage of examples relative to task instructions. \emph{Input:} system prompt. \emph{Output:} binary indicator for coverage gaps.
\end{itemize}

\paragraph{(iii) Output-Prompt Compliance (three tools).}
These tools validate generated outputs against requirements in the system prompt.
\begin{itemize}
\item \textbf{Output-Schema-Non-compliance-Detector} (LLM-based): checks that the LLM response satisfies the specified schema. \emph{Input:} system prompt, LLM output. \emph{Output:} binary violation indicator.
\item \textbf{Output-Instructions-Non-compliance-Detector} (LLM-based): verifies compliance of the response to prompt instructions. \emph{Input:} system prompt, LLM output. \emph{Output:} binary violation indicator.
\item \textbf{Output-Format-Violation-Detector} (LLM-based): ensures the response follows required output formatting. \emph{Input:} system prompt, LLM output. \emph{Output:} binary violation indicator.
\end{itemize}

\paragraph{(iv) Token-Level Anomalies in Outputs.}
Both are rule-based.
\begin{itemize}
\item \textbf{Unusual-Token-Detector} (rule-based): identifies non-standard characters and encoding issues. \emph{Input:} LLM output. \emph{Output:} binary indicator for unusual tokens.
\item \textbf{Token-Repetition-Detector} (rule-based): flags excessive repetition patterns (e.g., delimiters, escape sequences). \emph{Input:} LLM output. \emph{Output:} binary indicator for excessive repetition.
\end{itemize}

\paragraph{(v) Code-Specific Validation for Python.}
This tool is rule-based and applied selectively to code-bearing outputs.
\begin{itemize}
\item \textbf{Python-Code-Syntax-Validator} (rule-based): validates syntactic correctness and basic structural integrity using Python’s AST parser plus regex patterns for structural issues not captured by parsing. Applied only when the preceding detector is positive. \emph{Input:} LLM output. \emph{Output:} binary indicator for syntax errors.
\end{itemize}

\paragraph{(vi) Cross-Stage Information Consistency.}
\begin{itemize}
\item \textbf{Information-Consistency-Validator} (rule-based): detects factual inconsistencies between provided inputs and generated outputs, including truncation, modification, or hallucination of explicitly provided information. Implements regex patterns and fuzzy string matching for deterministic comparisons, and returns an inconsistency indicator alongside a detailed description of mismatched attributes and values. \emph{Input:} system prompt, user input, LLM output. \emph{Output:} binary indicator plus mismatch report.
\item \textbf{Reasoning-Action-Mismatch-Detector} (LLM-based): identifies discrepancies between the stated reasoning and final actions or conclusions in the response, reporting a binary mismatch indicator and a continuous severity score in $[0.0, 1.0]$. \emph{Input:} LLM output. \emph{Output:} binary indicator and severity score.
\end{itemize}

\paragraph{Python-Code-Segment-Detector.}
To ensure Python syntax validation is applied only to relevant agent outputs, we developed a detector that identifies Python code segments within LLM-generated text using weighted pattern recognition with conservative thresholds. The tool employs regular expressions and context filters to exclude JSON and prose content, prioritizing precision over recall to minimize false positives. It outputs a binary classification that determines whether responses should undergo downstream Python-specific analysis.

\paragraph{LLM-as-a-Judge Outputs.}
For the LLM-based tools, the evaluation includes detailed reasoning explanations and three actionable correction recommendations. This provides semantic analysis and targeted guidance, complementing the deterministic checks produced by the rule-based tools.

\section{Root Cause Analysis Tools for Multi-Agent LLM Systems}

The increasing deployment of multi-agent Large Language Model (LLM) systems in enterprise environments has introduced new challenges in system reliability and failure diagnosis. These agentic architectures, characterized by complex orchestration patterns involving multiple specialized agents with distinct roles and interdependencies, present unique failure modes that traditional debugging approaches cannot adequately address. Unlike monolithic LLM applications where failures are typically isolated to a single model interaction, multi-agent systems exhibit cascading failure patterns where errors in upstream agents propagate through the workflow, potentially amplifying issues or creating entirely new failure scenarios. The identification and classification of root causes in such systems is critical for maintaining system reliability, as a single misconfigured agent or poorly handled edge case can compromise the entire workflow's effectiveness. Furthermore, the heterogeneous nature of these systems—incorporating various validation mechanisms, code generation components, and reasoning modules—requires sophisticated analytical approaches that can operate across different failure dimensions simultaneously.

To address these challenges, we developed two complementary root cause analysis tools that operate at different levels of the system architecture. The first tool, an LLM-as-a-judge system, performs direct analysis of raw LLM call outputs within the workflow context. This tool employs advanced natural language processing techniques to detect execution failures, structural inconsistencies, and communication breakdowns between agents. It incorporates context-aware error detection algorithms that distinguish between actual system failures and benign patterns such as error-handling code or diagnostic comments, thereby minimizing false positive rates. The tool analyzes workflow sequences to identify deviations from expected agent interaction patterns while accommodating the inherent flexibility required in dynamic orchestration scenarios. Additionally, it provides detailed failure classification across multiple dimensions including structural integrity, execution success, communication effectiveness, and resource utilization, enabling systematic identification of the primary failure modes affecting system performance.

The second tool operates at a higher abstraction level, analyzing the aggregated outputs from multiple specialized validation tools rather than raw LLM responses. This meta-analytical approach processes results from nine distinct validation components, including consistency checkers, edge case analyzers, few-shot coverage evaluators, schema validators, and reasoning-action alignment detectors. Given the high failure detection rates observed in production systems—may be exceeding 80\% of LLM calls in complex workflows—this tool implements sophisticated aggregation and clustering strategies to manage information density effectively. The system employs hierarchical issue prioritization, grouping failures by severity and type, and presents findings through agent-level summarization and pattern detection algorithms. This approach transforms potentially overwhelming volumes of validation results into actionable insights by identifying recurring failure patterns, workflow bottlenecks, and systemic issues that span multiple agents. The tool's prompt engineering strategy dynamically adjusts information density based on failure volume, ensuring that critical issues receive detailed attention while maintaining analytical tractability even when processing hundreds of validation results per workflow execution.

\section{Case Study: IBM CUGA on AppWorld}
Having established our validation framework and root cause analysis methodology, we now evaluate their effectiveness through a comprehensive case study, applying them to execution logs from IBM CUGA across multiple LLM configurations and benchmark tasks. We applied the tools described above to the 24 tasks of AppWorld Benchmark. Three language models served as the agentic system engine:
\begin{itemize}
    \item \texttt{gpt-4o} solved 14 of 24 tasks (58.3\%).
    \item \texttt{mistral-medium-2505} solved 10 of 24 tasks (41.7\%).
    \item \texttt{llama-4-mav-17b-128e-instruct} solved 8 of 24 tasks (33.3\%).
\end{itemize}

In total, we recorded 1{,}940 LLM calls: 590 with \texttt{gpt-4o}, 411 with \texttt{mistral}, and 939 with \texttt{llama-4}. We omitted one out of 24 tasks from the Mistral analysis because it exhibited erratic behavior; its trace contained many LLM calls with empty inputs and outputs. After this exclusion, 336 Mistral calls remained.

For each LLM call, we analyzed the system prompts, user inputs, and LLM outputs using our tool framework. Unlike the other tools, \texttt{PythonCodeSyntaxChecker} was applied only to outputs were Python code was automatically detected.
In the context of IBM-CUGA system, each such llm-call was associated with an execution of an agent in the agentic workflow.
\begin{table}[h]
\centering
\resizebox{\linewidth}{!}{
\begin{tabular}{|l|p{0.65\linewidth}|}
\hline
\textbf{Tool} & \textbf{Issues} \\
\hline
LLMInputValidator & schema non-compliance, instruction non-compliance, format violations \\
\hline
LLMOutputValidator & schema non-compliance, instruction non-compliance, format violations \\
\hline
InformationConsistencyChecker & information inconsistency \\
\hline
EdgeCaseInstructionChecker & missing edge case instructions \\
\hline
FewShotCoverageChecker & coverage gaps \\
\hline
PromptConsistencyChecker & internal contradictions, example misalignment \\
\hline
ReasoningActionMismatchDetector & mismatch indicator \\
\hline
TokenAnomalyDetector & unusual token, excessive repetition \\
\hline
PythonCodeSyntaxChecker & syntax error \\
\hline
\end{tabular}}
\caption{Detected issues}
\label{tab:detected-issues}
\end{table}
Table~\ref{tab:detected-issues} summarizes the issues detected by each tool. 

\begin{figure}[ht] 
    \centering
    \includegraphics[width=\columnwidth]{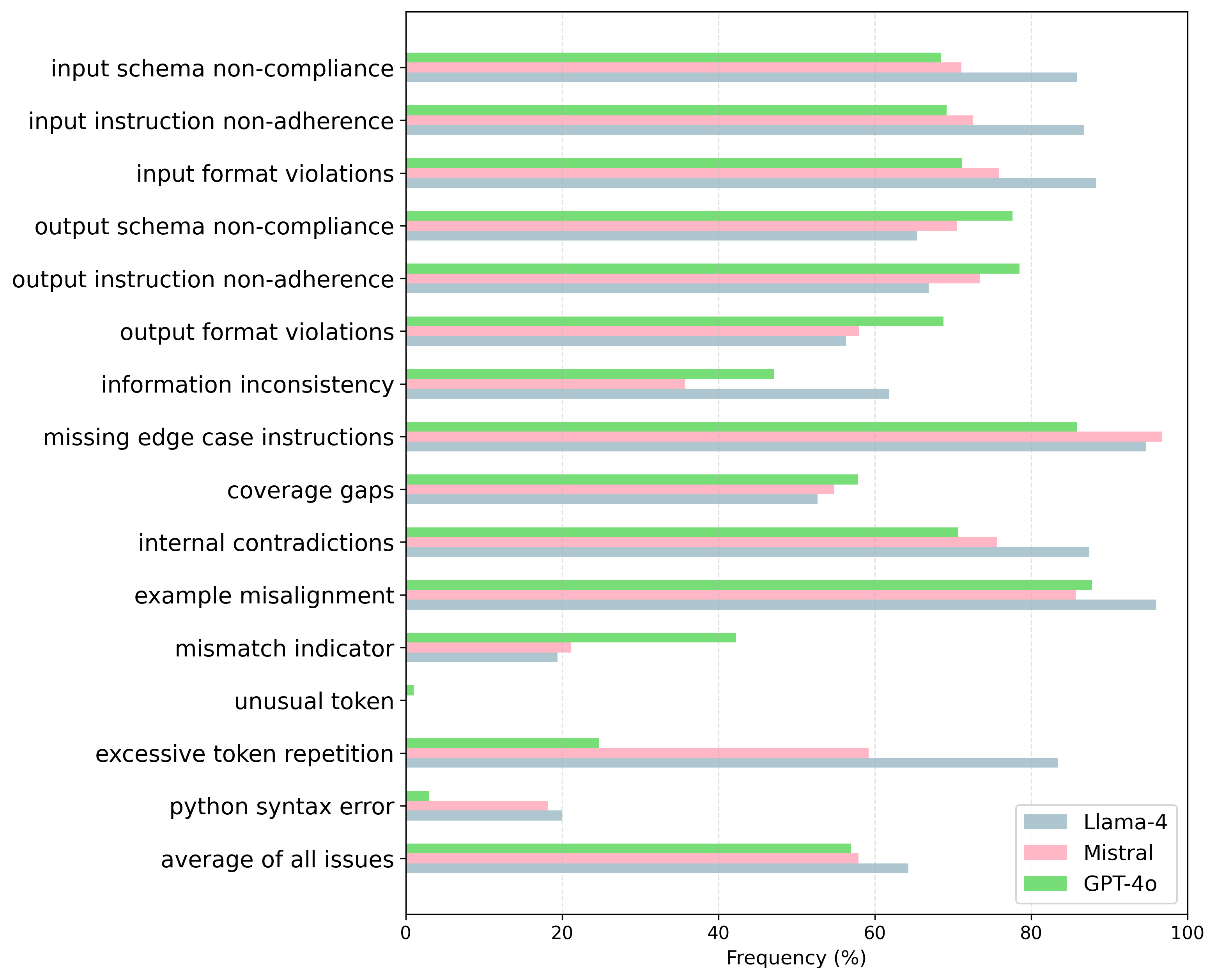}
    \caption{Issue frequency}
    \Description{Issue frequencies across LLM calls.}
    \label{fig:issue_frequency}
\end{figure}

Figure~\ref{fig:issue_frequency} shows issue frequencies across LLM calls. We observe numerous issues in input and output formats. It is important to note that our validation framework flags issues across three criticality levels: \textbf{Critical} (syntax errors, schema violations), \textbf{Moderate} (reasoning mismatches, format violations, information inconsistencies), and \textbf{Minor} (instruction adherence, coverage gaps, token anomalies). Many detected issues, particularly minor ones, do not cause execution failures as LLMs often self-correct small instruction violations or adapt to minor format deviations. However, addressing even minor issues systematically improves overall system reliability and reduces the cognitive load on downstream components. The high detection rates (64-88\%) reflect the sensitivity of our tools to any deviation from specified requirements, enabling proactive quality assurance rather than reactive debugging. We also observe that Python syntax errors are much less frequent for GPT-4o compared to the other two models.

\begin{table}[ht!]
\centering
\resizebox{\linewidth}{!}{
\begin{tabular}{l l c c}
\hline
\textbf{Model} & \textbf{Prompt Version} & \textbf{pass@3} & \textbf{pass@avg} \\
\hline
GPT\textendash4o & Original                               & 47\% & 41\% \\
                 & Plan+QA+Action                      & 50\% & 42\% \\
                 & Plan+QA+Action+Dec.        & 52\% & 45\% \\
\hline
LLaMA\textasciitilde4 Maverick 17B & Original             & 38\% & 32\% \\
                 & Plan+QA+Action                      & 42\% & 34\% \\
                 & Plan+QA+Action+Dec.        & 46\% & 35\% \\
\hline
Mistral Medium 2505 & Original                             & 35\% & 30\% \\
                 & Plan+QA+Action+Dec.        & 42\% & 34\% \\
\hline
\end{tabular}
}
\caption{Performance on WebArena GitLab subset under different prompt refinements. 
Models: GPT-4o (OpenAI), LLaMA~4 Maverick 17B, and Mistral Medium 2505.}
\label{tab:webarena-results}
\end{table}

%
Our agentic system comprises eight different agents. One of them, CodeAgent, is omitted from the analysis, as its relevance is primarily limited to Python code checks.

In several agents, including \textbf{Task Analyzer}, which initiates every trajectory, we observe frequent alerts regarding the input and output of LLM calls and their consistency with system prompt information. For example, the system prompt for \textit{API Planner }requires a summary encapsulating key observations from the input, but in some cases this summary is missing in the actual input. In contrast to most other agents, \textbf{Task Decomposition} exhibits a low percentage of input and output issues.

Few-shot coverage gaps are frequent for the two planning agents,\textbf{ API Planner} and \textbf{Plan Controller}, but are rare for the other agents. An example of reasoning behind a Plan Controller alert for this issue is: \textit{The examples provided do not cover scenarios where subtasks are partially completed with some subtasks in-progress and others not-started. Additionally, there is no example showing how to handle loops or when the planner needs to go back to a previous page.}

Edge case analysis appears to be a weak spot across all agents in the system, with a very high percentage of detected issues. An example of the reasoning provided by the corresponding tool for the \textbf{Shortlister} agent is: \textit{The prompt lacks explicit instructions for handling critical edge cases such as empty or missing inputs, invalid data formats, and ambiguous situations. It primarily focuses on the main task flow without addressing potential problems or errors that could arise during execution.}

\textbf{API Planner} stands out as the agent generating a large number of issues across most issue categories.

\subsection{Root cause tools}
We applied each of the two root cause tools on each of the model types and their respective failure detection tool results. For each log, each tool returned the root cause of a failure, if exists, along with a description of the affected agents, categorized failure severity (minor, moderate and critical) and remediation recommendation.
These are two examples demonstrating each of the root cause tool responses, both tagged as "critical":

\begin{itemize}
    \item \textbf{LLM-RC}: root cause tool has detected the following issue: "The APICodePlannerAgent and CodeAgent both failed to execute the task of calculating the total cost of items in the Amazon cart, citing a lack of necessary tools." where the described root cause was due to ".. lack of access to the necessary APIs and tools to perform the task". After a careful investigation of the log we discovered such a failed agent execution. This execution allowed for significant time saving in detecting the issue, as the log was ~200Mb in size.
    
    \item \textbf{T-RC}: root cause tool has detected the following issue: "The primary issue appears to be incorrect field names in JSON inputs, specifically the use of 'applications' instead of 'available\_apps', which leads to schema violations.". This was made easier as one of the inputs to the tool was the output of LLMInputValidator which checks schema violations between the requirements in the system prompt and the inserted information by the user. 
\end{itemize}

\begin{table*}[ht!]
\centering
\small
\begin{tabular}{p{4.5cm} p{2.5cm} p{4.6cm} p{4.6cm}}
\hline
\textbf{Task} & \textbf{Comparison} & \textbf{Single-Trace Tool: Root Cause} & \textbf{Comparison Tool: Root Cause} \\
\hline
Based on the question I posted about my last t-shirt order on amazon, has anyone experienced color fading after the first wash? Say yes or no
& Successful: GPT-4o, Failed: Mistral Medium 2505 & 
The APICodePlannerAgent did not produce any output, indicating a failure to generate the necessary pseudo-natural language plans for the CodeAgent & 
The APIPlannerAgent in the Failed Trace incorrectly identified the task as checking product reviews instead of responses to a specific question. \\
\hline
How many days of spotify premium subscription do I still have left? Round to the nearest number.
 & Successful: GPT-4o, Failed: LlaMA-4 Maverick 17B & 
The FinalAnswerAgent failed to synthesize the final output correctly, as it did not provide a valid final answer. & 
The CodeAgent in the Failed Trace miscalculated the remaining days due to incorrect rounding logic.
\\
\hline
\end{tabular}
\caption{Single-trace versus comparison tool; Examples of root-cause analysis}
\label{tab:comparison-examples}
\end{table*}

\subsection{Trace comparison tool}
%
%
In the previous sections, we described tools that analyze individual trajectories or aggregate results across runs, as well as separate tools for root-cause analysis and failure localization. We now consider an additional, complementary analysis mode. The same task may succeed when an agentic system uses one LLM yet fail with another—for example, succeeding with a frontier model but failing with an open-source alternative. Comparing successful and failed trajectories in such cases can reveal the underlying causes of failure.

%
To evaluate this hypothesis, we developed and compared two tools that simultaneously aim to locate failures and identify their root causes. The \textit{single-trace tool} analyzes the aggregated outputs of individual steps from the failed trajectory. A frontier model, \texttt{gpt-4o}, is used as an LLM-as-a-judge and is tasked with detecting the failure location, identifying root causes, and proposing recommendations for system improvement.

The \textit{comparison tool} analyzes both successful and failed trajectories. Using a prompt similar to that of the single-trace tool, it is required to produce the same diagnostic outputs and, additionally, generate a concise list of the main differences between the two trajectories.

%
Among the 24 tasks in the AppWorld Benchmark, 13 were successfully solved by at least one of the three evaluated models while at least one model failed on the same task. This yields 26 pairs of successful and failed trajectories. Both tools were applied to each pair, and their outputs were compared through manual inspection.

We observed that for 12 pairs (46.2\%), the comparison method provided better failure-location and root-cause explanations. For the remaining 14 pairs, the explanations were similar in quality, and in no case was the single-trace explanation superior.

%
Table \ref{tab:comparison-examples} presents two examples in which the comparison tool provided substantially better root-cause explanations. In both cases, the single-trace tool produced very general descriptions and failed to identify the underlying problem. In contrast, the comparison tool successfully pinpointed the actual root cause.

This small experiment demonstrates the strong potential of trace comparison tools, which, in our view, can be further enhanced beyond the brute-force version presented here.

\section{Validation-Informed System Improvements on WebArena}
To validate the practical utility of incidents discovered by our validation suite and the remediation strategies recommended by our root cause analysis tools, we implemented targeted improvements addressing similar failure patterns and evaluated their impact on IBM CUGA's performance in the WebArena benchmark. This evaluation demonstrates the actionable value of our diagnostic framework across different deployment contexts.
Based on systematic analysis of the identified error categories, prompt modifications were introduced in four stages:
\begin{itemize}
\item Planner: schema anchoring to reduce parsing errors.
\item QA: checks to prevent omissions or hallucinations.
\item Action: grounding of execution steps in API schemas.
\item Decomposition: structured few-shot breakdowns for complex tasks, with both general and web-specific variants.
\end{itemize}

In practice, several concrete fixes were implemented. We revised few-shot exemplars that contradicted core task guidelines and aligned them with consistent reasoning logic. We also standardized variable names and parameter references across user and system prompts to prevent mismatched argument bindings and parsing inconsistencies. These lightweight yet systematic adjustments improved prompt coherence and reduced propagation of minor formatting errors.

\subsection{Comparative Evaluation}
We assessed the effect of prompt refinements on the \textbf{WebArena GitLab subset} (204 samples), comparing models that differ markedly in scale and deployment trade-offs. Accuracy was measured using pass@$3$ and pass@avg, with results shown in Table~\ref{tab:webarena-results}. The evaluation included three representative models:  
\textbf{GPT-4o} (OpenAI) serves as a closed-source frontier baseline with strong reasoning and multi-modal alignment.  
\textbf{LLaMA~4 Maverick 17B} is an open-source, parameter-efficient model designed to balance capability and inference cost.  
\textbf{Mistral Medium} (\texttt{mistral-medium-2505}) is a mid-sized model optimized for enterprise-grade latency and cost efficiency. This spectrum highlights the trade-off between capability and deployment practicality: while GPT-4o delivers the highest raw accuracy, validation-driven refinements enable the smaller LLaMA~4 and Mistral models to approach frontier-level performance at a fraction of the cost.

\subsection{Improvement Analysis}
To measure stability of improvements, we compared task-level outcomes before and after refinement \cite{dror2018hitchhiker}. As shown in Table~\ref{tab:improvements}, most successful cases were preserved, while a non-trivial fraction of previously failing tasks were recovered. Regression cases were rare.

\begin{figure*}[ht!]
    \centering
    \includegraphics[width=0.7\textwidth,height=0.34\textheight]{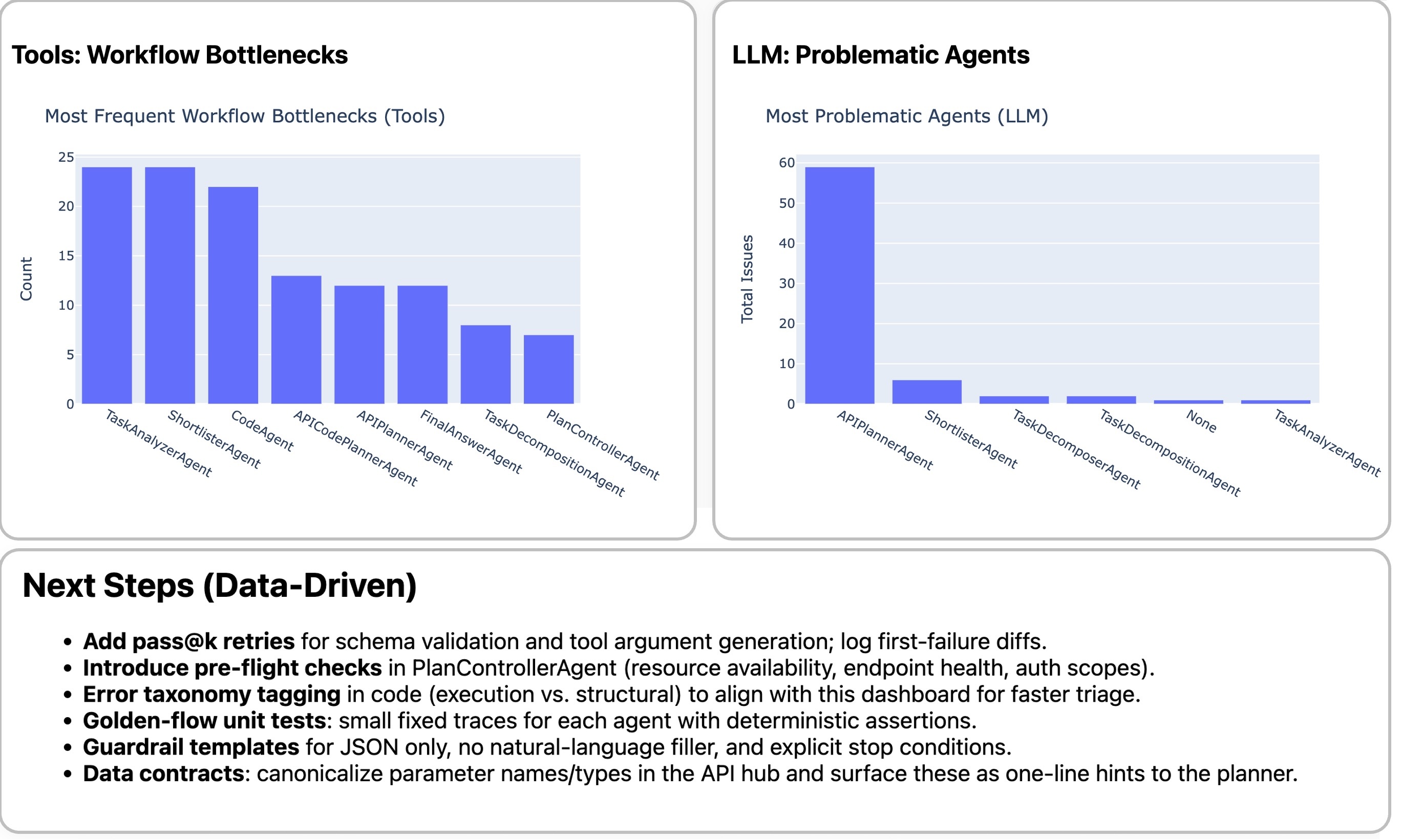}
    \caption{Interactive diagnostic dashboard for multi-agent system analysis. The left panel identifies workflow bottlenecks from tool-level logs, the right panel ranks agents by LLM reliability issues, and the bottom panel provides data-driven next steps for prompt refinements, structural redesigns, and cross-agent engineering practices.}
    \label{fig:dash5}
\end{figure*}

\subsection{Operational Takeaways}
These results show that validation-driven prompt refinement not only improves overall accuracy but also recovers failures that previously blocked execution. GPT performance improved modestly (47\% $\rightarrow$ 50\% at pass@3), but the larger gains came from smaller models: LLaMA~4 rose from 38\% to 42\%, and Mistral from 35\% to 42\%. Importantly, regression rates remained low, demonstrating robustness. 
From an operational perspective, these findings confirm that incident-guided design enables smaller, cost-efficient models to approach frontier-level accuracy. This reduces deployment cost, improves reliability, and supports safer scaling of agentic systems under enterprise constraints.

\begin{table}[h]
\centering
\resizebox{\linewidth}{!}{
\begin{tabular}{l c c c}
\hline
\textbf{Model} & \textbf{Preserved Success} & \textbf{Improved} & \textbf{Regressed} \\
\hline
GPT\textendash4o & 93 & 10 & 1 \\
LLaMA\textasciitilde4 Mav. & 72 & 12 & 4 \\
Mistral Med. & 74 & 8 & 2 \\
\hline
\end{tabular}}
\caption{Task-level improvement breakdown after prompt refinement (GitLab, 204 samples).}
\label{tab:improvements}
\end{table}

\section{Interactive Validation Analysis and Parsing Improvements}
To complement quantitative validation, we conducted an interactive experiment that transformed the framework's diagnostic outputs into an actionable decision process. We aggregated all validation reports-including root-cause analyses, few-shot examples, and parsing traces-along with each agent’s specification, and fed them into GPT-4o. The model engaged in a guided dialogue to identify dominant failure categories, rank agents by error severity, and recommend priority areas for refinement. This conversational analysis provided clear prioritization for improvement under time constraints, focusing particularly on agents with high rates of parsing and planner inconsistencies. An accompanying interactive HTML dashboard (see Figure \ref{fig:dash5}) visualized these findings, enabling rapid inspection of errors across agents and categories.

One key insight from this process was the prevalence of parsing-related failures, especially in smaller models such as \texttt{Llama-4} and \texttt{Mistral Medium}. Despite schema-constrained prompting, JSON parsing errors and malformed structures remained a major reliability bottleneck. In response, we implemented a \emph{sequential parsing mechanism} across all agents that attempts multiple parsing strategies in cascade-ranging from strict schema validation to heuristic recovery using fuzzy JSON repair and code execution feedback. This mechanism significantly reduced variability in parsing outcomes and improved overall execution robustness across model scales. Together, the conversational reflection and adaptive layer demonstrate how interactive validation can inform both strategic prioritization and concrete architectural improvements in agentic systems.

\section{Discussion and Conclusions}

\subsection{Lessons Learned for Agentic Engineering}
Our study highlights three overarching lessons for the engineering of reliable agentic systems.  
First, validation must be treated as a continuous engineering process rather than an evaluation afterthought. Agentic systems evolve as their environments, tools, and foundation models change; maintaining reliability therefore requires feedback loops that connect incident detection, diagnosis, and design refinement.  
Second, schema and interface alignment are critical to sustaining autonomy across heterogeneous components. Even when language models achieve strong reasoning performance, small inconsistencies between prompts, schemas, or APIs remain a dominant source of failure. Automated validation and adaptive parsing strategies can mitigate these risks by enforcing structural consistency without constraining model creativity.  

Third, mixed-initiative validation-where models reflect on their own outputs-opens new design opportunities. Feeding diagnostic traces, root causes, and system descriptions back into an LLM enabled self-reflection that prioritized where to intervene first. This approach illustrates a broader principle: validation can itself become agentic, combining automated analysis with human oversight to accelerate improvement.  
Across these findings, we observe that agentic engineering increasingly resembles traditional software engineering disciplines, requiring architectural abstractions, testing frameworks, and observability stacks, while introducing new challenges around semantic correctness and runtime adaptability.

\subsection{Implications and Future Directions}
The broader implication is that dependability in agentic systems can be engineered through disciplined process, monitoring, and design, not merely through larger models. Smaller or domain-specific models can deliver high reliability when paired with structured validation, adaptive parsing, and iterative refinement.
For the agentic engineering community, this points to a research agenda that unifies AgentOps practices with software-engineering principles: runtime validation, failure taxonomies, tool-agnostic observability, and self-improving architectures. Future work should generalize these methods beyond individual frameworks, integrating validation, testing, and reflection into standard development lifecycles.  
In summary, the lessons from our validation framework suggest that the path to robust agentic systems lies in treating them as continuously engineered software ecosystems. Embedding quality assurance, interpretability, and adaptation directly into their operation moves the field toward reliable, transparent, and scalable agentic systems-an essential step for the emerging discipline of agentic engineering.


\bibliographystyle{ACM-Reference-Format}
\bibliography{bib}

\end{document}